\documentclass[10pt,conference,a4paper]{IEEEtran}

\usepackage{graphicx}
\usepackage{caption}
\usepackage{subcaption}
\usepackage{amsmath}
\usepackage{algorithmic}
\usepackage{algorithm}
\usepackage{lineno}
\usepackage{multirow}
\usepackage{tabularx}
\usepackage{booktabs}
\usepackage{url}
\usepackage{color}
\usepackage[figuresright]{rotating}
\usepackage[utf8]{inputenc}
\usepackage[T1]{fontenc}

\newcolumntype{Y}{>{\arraybackslash}X}

\begin{document}

\title{Two-Stage Resampling for Convolutional Neural Network Training in the Imbalanced Colorectal Cancer Image Classification}

\author{\IEEEauthorblockN{Michał Koziarski}
\IEEEauthorblockA{Department of Electronics\\
AGH University of Science and Technology\\
Al. Mickiewicza 30, 30-059 Kraków, Poland\\
Email: michal.koziarski@agh.edu.pl}}

\maketitle

\begin{abstract}
Data imbalance remains one of the open challenges in the contemporary machine learning. It is especially prevalent in case of medical data, such as histopathological images. Traditional data-level approaches for dealing with data imbalance are ill-suited for image data: oversampling methods such as SMOTE and its derivatives lead to creation of unrealistic synthetic observations, whereas undersampling reduces the amount of available data, critical for successful training of convolutional neural networks. To alleviate the problems associated with over- and undersampling we propose a novel two-stage resampling methodology, in which we initially use the oversampling techniques in the image space to leverage a large amount of data for training of a convolutional neural network, and afterwards apply undersampling in the feature space to fine-tune the last layers of the network. Experiments conducted on a colorectal cancer image dataset indicate the usefulness of the proposed approach.
\end{abstract}

\IEEEpeerreviewmaketitle

\section{Introduction}

Data imbalance occurs in the classification task whenever one of the classes (\textit{minority class}) consists of a smaller number of observations than one of the other classes (\textit{majority class}). Data imbalance poses a significant challenge for the traditional learning algorithms, a majority of which was designed with the assumption of balanced class distributions. When trained on the imbalanced data, such algorithms tend to display a bias towards the majority class at the expense of the minority class, degrading the overall performance. Due to its prevalence in real-life applications and a potentially significant impact on the training process of the classification algorithms, imbalanced data learning is a well-established area of machine learning. A large body of techniques, which can be divided into the \textit{data-level} and \textit{algorithm-level} approaches, exists in the literature. The aim of the former is modification of the training data prior to classification to balance the class distribution, either by removing some of the existing majority class observations (\textit{undersampling}) or by creating new minority class observations (\textit{oversampling}). The latter type of methods directly modifies the training process of the classifier to compensate for the imbalanced data distribution. However, despite the large body of existing work, a majority of data-level approaches translates poorly to the image data. More sophisticated oversampling techniques, such as SMOTE \cite{chawla2002smote}, are based on the notion of generating synthetic minority objects by interpolating a pair of existing, neighboring observations. When applied directly to the image data, which became a de facto standard for model input with the advent of deep learning, this can produce unrealistic images with visible artifacts. At the same time, SMOTE is a cornerstone for the contemporary data-level approaches for handling data imbalance \cite{fernandez2018smote}, with a majority of existing oversampling approaches based directly on the idea of synthetic oversampling by interpolation. Similarly, applying undersampling can also be problematic when training deep convolutional neural networks, as they usually require a large amount of training data to achieve a satisfactory performance. As a result, one of the challenges in the imbalanced image recognition domain is adapting the existing research to be applicable within the convolutional neural network training.

In this paper we propose a novel approach for handling data imbalance in the image recognition task, in which we apply data resampling in two stages: first of all we oversample the data directly in the image space and use it for the initial training of the model, and afterwards we undersample the data in the high-level feature space produced, based on the input images, by the previously trained network, to fine-tune its last layers. Afterwards we experimentally evaluate the proposed approach on a colorectal cancer image datatset. The rest of the paper is organized as follows. In Section~\ref{sec:rl} we present the related work on the subject of imbalanced image recognition, with particular focus on histological image classification. In Section~\ref{sec:met} we discuss the shortcomings of the existing data-level approaches in more detail and present our two-stage resampling approach. In Section~\ref{sec:exp} we present the details of the conducted experimental study, including the details of the used dataset and the data imbalance artificially introduced into it. Finally, in Section~\ref{sec:conc} we present our conclusions.

\section{Related Work}
\label{sec:rl}

First work related to the subject of the impact of data imbalance on neural networks can be traced to Anand et al. \cite{anand1993improved}, who considered the case of shallow neural networks. More recently, Johnson and Khoshgoftaar \cite{johnson2019survey} conducted a survey on deep learning with class imbalance, outlining the recent advances in the field, in particular in the context of convolutional neural networks. Some of the most relevant approaches outlined in their study include the works by Masko and Hensman \cite{masko2015impact}, who used random oversampling (ROS) before training the convolutional neural network, Lee et al. \cite{lee2016plankton}, who use random undersampling (RUS) in combination with transfer learning, Pouyanfar et al. \cite{pouyanfar2018dynamic}, who introduced a dynamic sampling approach, and Buda et al. \cite{buda2018systematic}, who consider the impact of RUS, ROS, and two-phase training, an approach in which the network is first trained on the balanced dataset, and afterwards finetuned on the original data. Conceptually, the last approach is the most similar to the one presented in this paper. However, the method we advocate for uses resampling in both the image space and the feature space, whereas two-phase training uses only the former.

In the context of the histological image classification, Koziarski et al. \cite{koziarski2018convolutional} considered the impact of data imbalance on the performance of a convolutional neural network in the breast cancer recognition task, as well as the possibility of applying different data resampling techniques directly in the image space. Later on, Koziarski \cite{koziarski2019radial} successfully applied the Radial-Based Undersampling algorithm in the high-level feature space, achieving an improvement in the performance. Additionally, Kwolek et al. \cite{kwolek2019breast} considered the possibility of applying the active learning in the same problem domain.

Finally, it is important to note that besides the attempts at translating the traditional data-level approaches to the image recognition setting, methods dedicated to image data and convolutional neural network training can be found in the literature. Examples of such methods include BAGAN \cite{mariani2018bagan}, a generative adversarial network applied to the image synthesis task, and the class-balanced loss function \cite{cui2019class}.

\section{Two-Stage Resampling}
\label{sec:met}

\begin{figure}
  \centering
   \includegraphics[width=\linewidth]{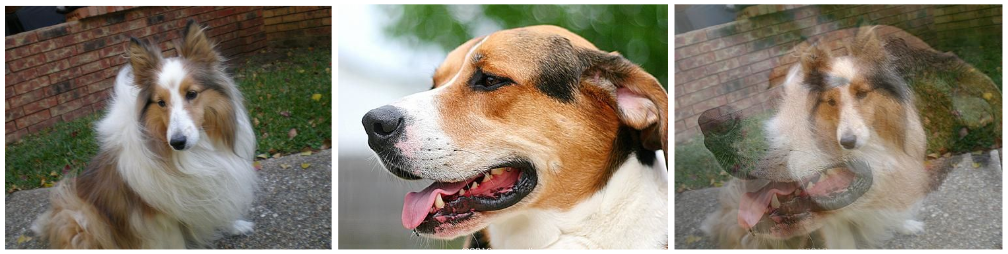}
  \caption{An example of SMOTE interpolation applied to a high-dimensional image data.}
  \label{fig:smote}
 \end{figure}

Both the over- and the undersampling techniques have their limitations that have to be addressed to achieve a satisfactory performance, in particular in the image recognition setting. Traditional oversampling algorithms producing synthetic observations, such as SMOTE and its derivatives, were not designed with the intention of being used on the image data. They are based on the idea of using data interpolation between nearby minority observations to synthesize new ones, and when applied in the image space, this approach can lead to creation of visual artifacts. As a result, produced synthetic images can be unrealistic, in particular when the considered data is highly dimensional. This point was illustrated in Figure~\ref{fig:smote}. On the other hand, strategies that duplicate the existing observations, namely random oversampling, are known to cause overfitting in the simpler classification algorithms, such as decision trees \cite{chawla2002smote}. Conceptually, this effect is likely to be even further pronounced for the convolutional neural networks, which tend to have a significantly higher number of trainable parameters. Finally, undersampling strategies reduce the amount of data available for training, which is crucial for a successful training of large neural networks.

\begin{figure*}
\centering
\includegraphics[width=\textwidth]{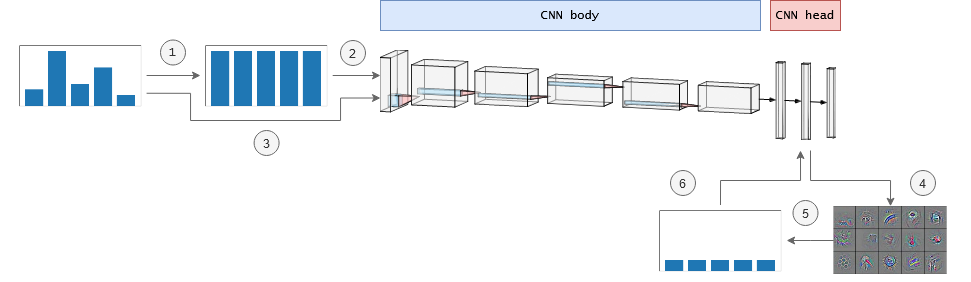}
\caption{Visualization of the data flow in the proposed approach. In the first stage, original image data is oversampled (1) and used to train the convolutional neural network (2). In the second stage, image data is passed through the trained convolutional neural network (3) and high-level features are extracted (4). These features are afterwards undersampled (5) and used to fine-tune the networks head (6).}
\label{fig:diagram}
\end{figure*}

To address the aforementioned shortcomings of the over- and undersampling techniques, namely the possibility of overfitting in the case of random oversampling, the introduction of visual artifacts in a high-dimensional synthetic observations generated by SMOTE-based algorithms, and the loss of data in the case of undersampling, we propose a conceptually simple strategy of two-stage data resampling intertwined with a traditional convolutional neural network training procedure. The motivation behind the approach is to, first of all, leverage a high amount of data, further enhanced by applying oversampling, in the initial training of the convolutional network, and afterwards fine-tune the fully-connected head of the network on a smaller amount of undersampled data, uncontaminated by the synthetic observations. We present a visualization of the data flow of the proposed approach in Figure~\ref{fig:diagram}.

It is important to note that even though we propose using oversampling in the first stage of the algorithm and undersampling in the second stage, the exact choice of both resampling strategies can be treated as a parameter of the method. Another theoretically viable strategy is applying one of the variants of SMOTE in the second resampling stage, directly on the high-level features extracted from a previously trained convolutional network. In principle, such approach could reduce the impact of visual artifacts, introduced if SMOTE was used in the image space, on the network training process. Secondly, in particular the first resampling stage could be omitted entirely. The essential steps of the method are, first of all, initial training of the network, and secondly, using the trained network to extract high-level features, afterwards resampled and utilized during fine-tuning of the fully-connected head of the network. As a final remark, while we propose performing second stage resampling directly in the high-level feature space, it is worth mentioning that the order of feature extraction and resampling is relevant only for the non-random approaches. During both the random over- and undersampling, performing the resampling prior to feature extraction would produce an identical result. The order of the operations is, however, vital when using guided resampling strategies, such as SMOTE.

\section{Experimental Study}
\label{sec:exp}

\subsection{Dataset}

\begin{figure}
\centering
\includegraphics[width=\linewidth]{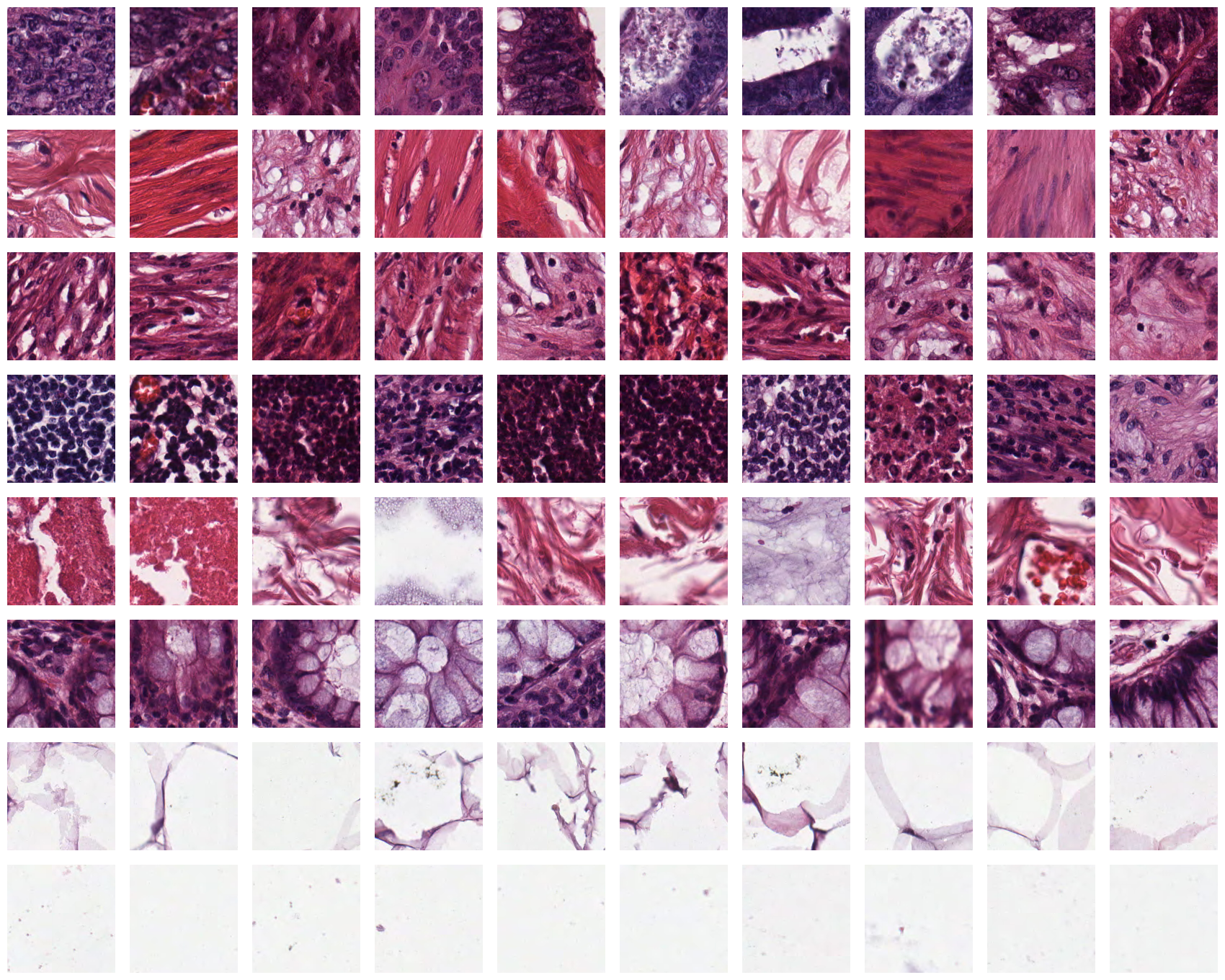}
\caption{Sample images from the used dataset, with individual rows containing samples from a single class. From the top: (1) tumour epithelium, (2) simple stroma, (3) complex stroma, (4) immune cell conglomerates, (5) debris and mucus, (6) mucosal glands, (7) adipose tissue, (8) background.}
\label{fig:samples}
\end{figure}

Despite the prevalence of data imbalance in real-life image recognition, a majority of popular benchmark datasets consists of data with completely balanced distribution. To the best of our knowledge no dedicated image recognition benchmarks for evaluation of the impact of data imbalance exist. Because of that, in this paper we considered the case of an originally balanced benchmark dataset with artificially introduced data imbalance. Specifically, we used a colorectal cancer histology dataset published by Kather et al. \cite{kather2016multi}. It consists of a total of 5,000 histological images of human colorectal cancer divided into eight different types of tissue. The dataset included textures extracted at different scales, from individual cells to larger structures. Each image had a dimensionality of $150\times150$ pixels. Sample images from the dataset were presented in Figure~\ref{fig:samples}.

\begin{figure*}
\centering
\includegraphics[width=0.75\textwidth]{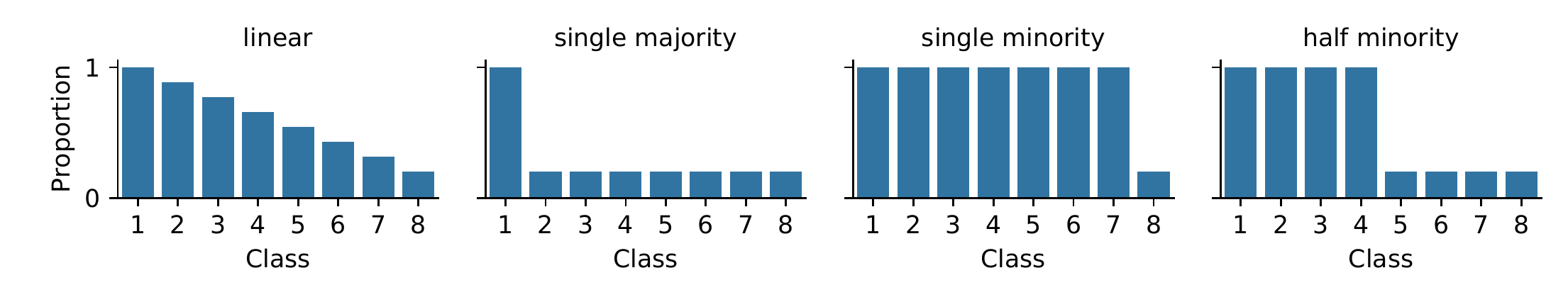}
\caption{Class distribution after introducing data imbalance to the dataset in each of the considered imbalance scenarios.}
\label{fig:proportion}
\end{figure*}

Data imbalance was introduced by randomly undersampling the original data up to the point of achieving the desired imbalance ratio (IR). 4 different imbalance levels were considered: balanced (IR = 1.0), small (IR = 2.0), medium (IR = 5.0) and high (IR = 10.0), as well as 4 different imbalance types: linear, single majority, single minority and half minority. Used imbalance types were chosen with the aim of simulating different types of relations between the classes that can occur in the multi-class setting. In the case of linear imbalance, the ratio of observations for $i$-th class was calculated as
\begin{equation}
    r_i = 1 + \frac{\textrm{IR} - 1}{M - 1} \cdot (i - 1),
\end{equation}
with $M = 8$ denoting the number of classes, and $i \in \{1, 2, ..., 8\}$, whereas for the three remaining imbalance types it was equal to
\begin{equation}
r_i = 
\begin{cases}
    \textrm{IR}, & \text{if } i\text{-th class was a minority class}\\
    1,              & \text{otherwise}
\end{cases}
\end{equation}
Finally, the desired number of observations for $i$-th class was calculated as
\begin{equation}
    n_i = \frac{1}{r_i} \cdot n,
\end{equation}
with $n = 625$ denoting the total number of observations per class. Class distributions for different imbalance types were presented in Figure~\ref{fig:proportion}. 

To reduce the impact of random data variability on the results, 10-fold cross-validation was used to partition the original dataset. Afterwards, for each of the folds classes were randomly ordered, and the undersampling was performed iteratively for the increasing imbalance levels. In other words, for a given fold, the assignment of a class to either the majority or the minority was preserved across the imbalance levels and types, and the dataset for a given imbalance level was always a subset of a dataset for a lower imbalance level for the same fold.

\begin{figure}
\centering
\includegraphics[width=\linewidth]{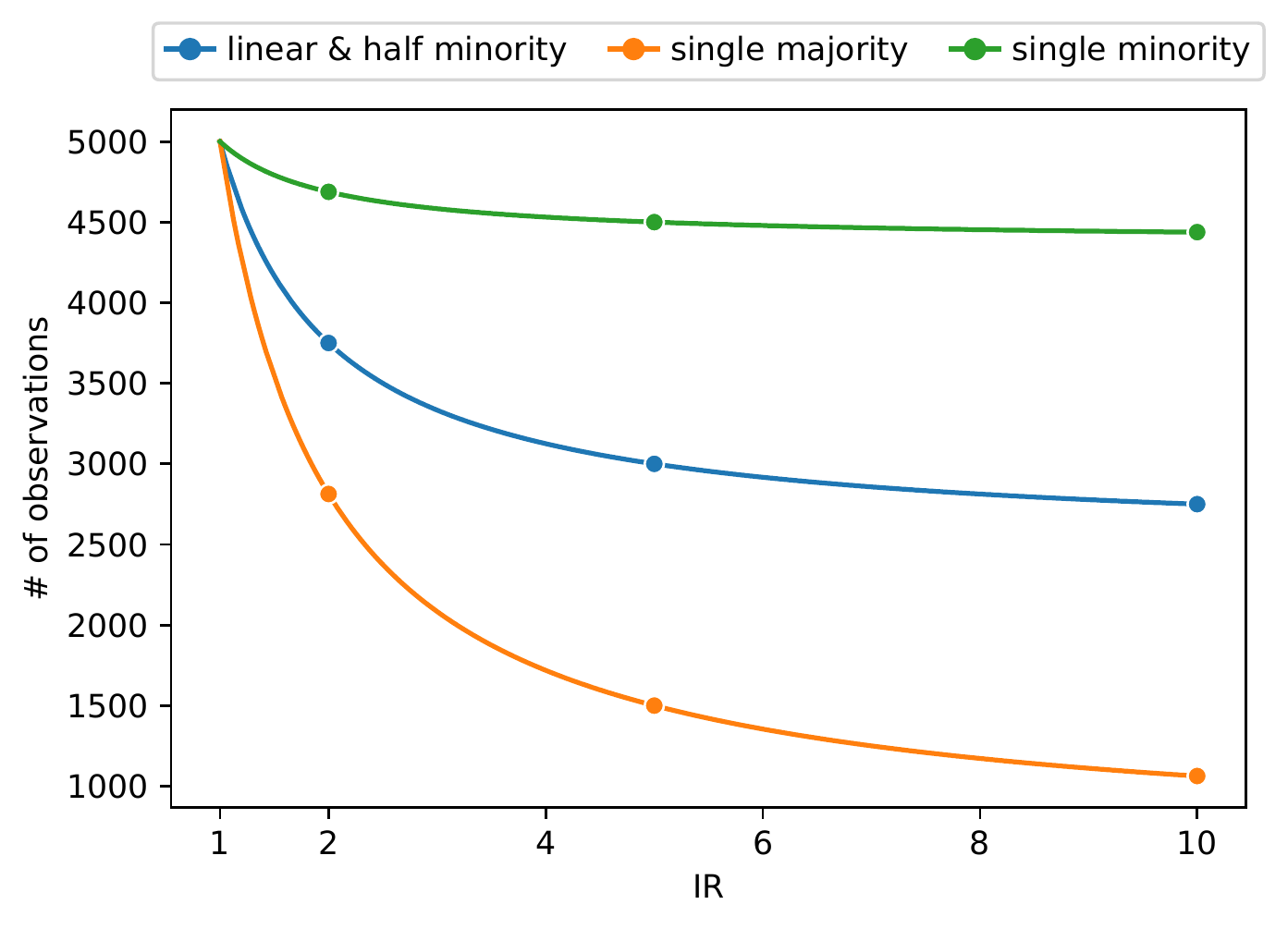}
\caption{The impact of the choice of an imbalance scenario on the relationship between the imbalance ratio (IR) and the total number of observations.}
\label{fig:n_obs}
\end{figure}

In total, 160 data partitionings, with different combinations of imbalance levels, imbalance types, and folds, were created. They represented different levels of severity of data imbalance in the dataset. It is worth mentioning that the imbalance type affected the associated difficulty in two ways. It not only altered the number of minority classes, but also impacted the total amount of data available for training. This was illustrated in Figure~\ref{fig:n_obs}. As can be seen, applying the single majority scenario led to a lowest total number of observations, whereas the single minority resulted in the opposite.

\subsection{Set-up}

\noindent \textbf{Classification.} We based our experimental analysis on the MobileNet architecture \cite{howard2017mobilenets}, with the input shape adjusted to $150\times150$, the number of neurons in the only fully-connected layer adjusted to 1024, and the number of neurons in the softmax layer adjusted to 8. During the second resampling stage we passed the original images through the network and extracted 1024-dimensional feature representations. Afterwards, we resampled the data in the feature space and used it to fine-tune the fully-connected and softmax layers. In both stages we trained the network using RMSprop optimizer for 20 epochs, with learning rate equal to 0.0001, batch size equal to 32, categorical crossentropy used as the loss function, and the weights of the model pretrained on the ImageNet dataset \cite{deng2009imagenet}. We based our implementation on the Keras library \cite{chollet2015keras}.

\noindent \textbf{Resampling.} In addition to the baseline setting in which no data resampling was applied, we considered three popular data-level approaches for handling data imbalance: random undersampling (RUS), random oversampling (ROS) and SMOTE. The methods were applied in one of three ways: directly in the image space (IS), in which case the data was vectorized prior to resampling and reshaped to the original format afterwards; in the feature space (FS), in which case network was first trained on the imbalanced data and afterwards the fully-connected and softmax layers were fine-tuned on the resampled data; and in the two-stage setting (TS), in which case the data was first resampled in the image space, used to train the network, and afterwards a high-level feature representations were extracted from the original data, resampled with a different algorithm, and used to fine-tune the fully-connected and softmax layers. In all of the cases we performed resampling up to the point of achieving a balanced class distribution. In the case of SMOTE we used the number of nearest neighbors $k = 5$. Our implementation was based on the imbalanced-learn library \cite{lemaitre2017imbalanced}.

\noindent \textbf{Evaluation.} The metric most commonly used to evaluate the performance of image recognition algorithms is classification accuracy (Acc). However, it is not a reliable measure of performance when dealing with imbalanced data, since it assigns weight of the miss-classification of individual classes as proportional to the number of observations that they consist of. To alleviate this issue, in this paper we use three different metrics proposed explicitly for the multi-class imbalanced data setting: Average Accuracy (AvAcc), Class Balance Accuracy (CBA), and Geometric Average of Recall (MAvG) \cite{Branco:2017}. They can be defined as follows:
\begin{align}
AvAcc &= \frac{\sum_{i=1}^M TPR_i}{M},\\
CBA &= \frac{\sum_{i=1}^{M}\frac{mat_{i,i}}{max\left( \sum_{i=1}^{M} mat_{i,j},\sum_{i=1}^{M} mat_{j,i} \right)}}{M},\\
MAvG &= \sqrt[M]{\prod_{i=1}^M recall_i},
\end{align}
with $M$ denoting the number of classes, $TPR_i$ denoting the true positive rate for class $i$, and $mat_{i,j}$ denoting the number of instances of the true class $i$ that were predicted as class $j$.

\subsection{Results}

\begin{figure*}
\centering
\includegraphics[width=0.75\textwidth]{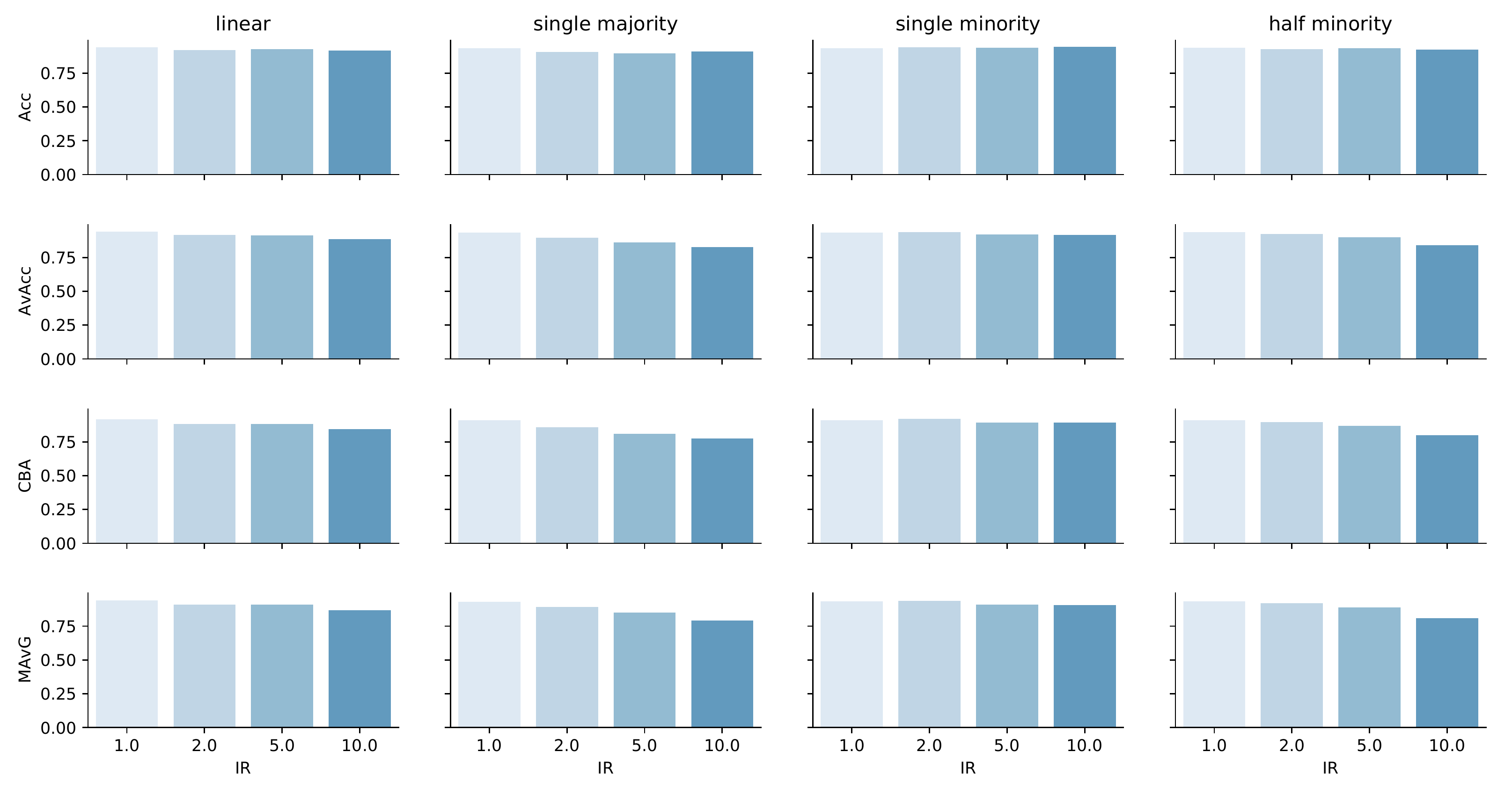}
\caption{The impact of an increasing imbalance ratio (IR) on the performance of a network trained without any data resampling, individually for each imbalance scenario and performance metric.}
\label{fig:ir-impact}
\end{figure*}

We began the conducted analysis with an experiment, in which we evaluated the impact of imbalance level on the classification performance when no data resampling was applied. We considered both the different imbalance scenarios as well as different metrics as a separate test cases. As a reference, we also included the classification accuracy. The results of this stage of the experimental analysis were presented in Figure~\ref{fig:ir-impact}. As can be seen, especially the higher levels of imbalance were highly detrimental to the classification performance with respect to the dedicated metrics for evaluation of performance in the multi-class imbalance setting. This was the case even though, depending on the type of considered imbalance scenario, performance drop was not visible, or was much smaller, with respect to the classification accuracy. This was shown to further emphasize the fact that the classification accuracy is not an appropriate metric in the imbalanced data classification task. Furthermore, it is worth mentioning that the degree of the performance drop was dependant on the specific imbalance scenario. In particular, in the case of linear and half minority scenarios, where the total number of observations available for training was the same, the degree of performance drop still differed: final performance was noticeably smaller in the half minority scenario. This illustrates the point that the decrease in the number of observations used for training was not the sole reason of the performance drop, and the nature of the imbalance relations between the classes is also an important factor.

In the second stage of the conducted experimental analysis we considered the impact of applying different resampling strategies directly in the image space. We used all of the considered resampling algorithms, namely RUS, ROS and SMOTE, and compared them with the baseline case in which no resampling was applied. The results, once again separated into specific imbalance scenarios and imbalance levels, were presented in Table~\ref{tab:baseline}. As can be seen, the choice of the resampling strategy leading to the best performance depended on the specific imbalance scenario, as well as the considered performance metric. However, several observations can be made based on the achieved results. Most importantly, using RUS in no case led to the best results. In fact, in the majority of the cases training on the undersampled data led to a worse performance than the baseline setting, in which no resampling was applied. On the other hand, using both ROS and SMOTE led to an improved performance in the majority of imbalance scenarios, with neither of the methods producing better performance in all of the settings. The only exception to this trend was the single minority scenario, based on the results presented in Figure~\ref{fig:ir-impact} the least severe type of imbalance, in which using neither of the oversampling strategies led to a performance better than the baseline case. Interestingly, this held true even for higher imbalance levels. The performance improvement due to oversampling was, in general, highest for the most difficult scenario, the single majority. Finally, it is worth noting that the fact that applying SMOTE directly in the image space produced a performance improvement is somehow surprising. As previously discussed, we attribute it to the characteristics of this particular dataset, for which relatively small dimensionality of an individual image, as well as the nature of a histological data, which consists of textures, led to a less severe visual artifacts due to the image interpolation. It is likely that this trend would not translate to a natural images with higher dimensionality, such as the ones comprised in the ImageNet.

\begin{table*}
\begin{center}
\caption{A comparison of different resampling strategies applied directly on the image data, individually for each imbalance scenario, imbalance ratio (IR) and performance metric.}
\label{tab:baseline}
\begin{tabularx}{\textwidth}{l @{\extracolsep{\fill}} llllll}
\toprule
Scenario & IR & Metric & Baseline & IS\textsubscript{RUS} & IS\textsubscript{ROS} & IS\textsubscript{SMOTE} \\
\midrule
linear & 2.0 & AvAcc & 0.9167 & 0.9114 & 0.9186 & \textbf{0.9238} \\
 &  & CBA & 0.8811 & 0.8737 & 0.8875 & \textbf{0.8940} \\
 &  & MAvG & 0.9100 & 0.9058 & 0.9136 & \textbf{0.9198} \\
\cmidrule{2-7}
 & 5.0 & AvAcc & 0.9133 & 0.8832 & 0.9215 & \textbf{0.9219} \\
 &  & CBA & 0.8832 & 0.8202 & 0.8882 & \textbf{0.8903} \\
 &  & MAvG & 0.9083 & 0.8756 & 0.9154 & \textbf{0.9162} \\
\cmidrule{2-7}
 & 10.0 & AvAcc & 0.8874 & 0.8603 & 0.8860 & \textbf{0.8979} \\
 &  & CBA & 0.8456 & 0.7769 & 0.8577 & \textbf{0.8655} \\
 &  & MAvG & 0.8676 & 0.8511 & 0.8635 & \textbf{0.8841} \\
\midrule
single majority & 2.0 & AvAcc & 0.8976 & 0.9071 & 0.9131 & \textbf{0.9167} \\
 &  & CBA & 0.8576 & 0.8731 & 0.8795 & \textbf{0.8851} \\
 &  & MAvG & 0.8915 & 0.9018 & 0.9090 & \textbf{0.9111} \\
\cmidrule{2-7}
 & 5.0 & AvAcc & 0.8633 & 0.8727 & 0.8781 & \textbf{0.8899} \\
 &  & CBA & 0.8093 & 0.8040 & 0.8253 & \textbf{0.8429} \\
 &  & MAvG & 0.8519 & 0.8645 & 0.8643 & \textbf{0.8792} \\
\cmidrule{2-7}
 & 10.0 & AvAcc & 0.8267 & 0.8333 & 0.8467 & \textbf{0.8476} \\
 &  & CBA & 0.7765 & 0.7323 & \textbf{0.7985} & 0.7842 \\
 &  & MAvG & 0.7907 & 0.7412 & 0.7589 & \textbf{0.8168} \\
\midrule
single minority & 2.0 & AvAcc & 0.9381 & 0.9159 & \textbf{0.9383} & 0.9327 \\
 &  & CBA & \textbf{0.9185} & 0.8818 & 0.9165 & 0.9076 \\
 &  & MAvG & 0.9355 & 0.9117 & \textbf{0.9357} & 0.9280 \\
\cmidrule{2-7}
 & 5.0 & AvAcc & \textbf{0.9202} & 0.8793 & 0.9183 & 0.9159 \\
 &  & CBA & 0.8910 & 0.8244 & \textbf{0.8949} & 0.8848 \\
 &  & MAvG & \textbf{0.9104} & 0.8707 & 0.9041 & 0.9047 \\
\cmidrule{2-7}
 & 10.0 & AvAcc & \textbf{0.9179} & 0.8573 & 0.8985 & 0.9057 \\
 &  & CBA & \textbf{0.8912} & 0.7684 & 0.8784 & 0.8803 \\
 &  & MAvG & \textbf{0.9068} & 0.8468 & 0.7963 & 0.8047 \\
\midrule
half minority & 2.0 & AvAcc & 0.9226 & 0.9101 & 0.9246 & \textbf{0.9254} \\
 &  & CBA & 0.8963 & 0.8630 & \textbf{0.8966} & 0.8925 \\
 &  & MAvG & 0.9193 & 0.9057 & 0.9211 & \textbf{0.9216} \\
\cmidrule{2-7}
 & 5.0 & AvAcc & 0.8993 & 0.8867 & 0.8776 & \textbf{0.9044} \\
 &  & CBA & 0.8670 & 0.8110 & 0.8467 & \textbf{0.8703} \\
 &  & MAvG & 0.8884 & 0.8798 & 0.8608 & \textbf{0.8979} \\
\cmidrule{2-7}
 & 10.0 & AvAcc & 0.8395 & 0.8534 & \textbf{0.8724} & 0.8540 \\
 &  & CBA & 0.7984 & 0.6752 & \textbf{0.8405} & 0.8165 \\
 &  & MAvG & 0.8089 & 0.8387 & \textbf{0.8440} & 0.8261 \\
\bottomrule
\end{tabularx}
\end{center}
\end{table*}

Finally, in the last stage of the conducted experiments we directly compared different approaches to applying resampling: in the image space (IS), high-level feature space (FS) and in the two-stage setting (TS). For both the IS and FS we evaluated all three of the considered resampling strategies: RUS, ROS and SMOTE. For the two-stage resampling, based on the results obtained for IS and FS, as a preferred resampler combination we used SMOTE in the first stage to oversample the data in the image space, and afterwards RUS to undersample high-level features for further fine-tuning. Additionally, as a reference we also considered the variant in which SMOTE was used in both resampling stages. For each of the considered approaches we evaluated three imbalance levels (IR $\in \{2.0, 5.0, 10.0\}$) and all four of the proposed imbalance scenarios, with 10-fold cross-validation used for every setting. In total, every strategy was evaluated on 120 different data partitionings. Averaged results, as well as the average ranks achieved by each method, were presented in Table~\ref{tab:final-summary}. As can be seen, using the proposed approach led to achieving the best performance with respect to all of the considered metrics, both with respect to the averaged scores as well as average ranks. Importantly, it also outperformed the two-stage resampling variant which used SMOTE exclusively, indicating the suitability of using undersampling in the high-level feature space.

\begin{table*}
\begin{center}
\caption{A comparison of image space (IS), high-level feature space (FS) and two-stage (TS) resampling, with the specific resampler used denoted in subscript. In addition to the scores averaged over all imbalance scenarios, average ranks achieved by each method were presented in the parentheses.}
\label{tab:final-summary}
\begin{tabularx}{0.75\textwidth}{l @{\extracolsep{\fill}} lll}
\toprule
Approach & AvAcc & CBA & MAvG \\
\midrule
Baseline & 0.8952 (6.14) & 0.8596 (6.31) & 0.8824 (6.14) \\
IS\textsubscript{RUS} & 0.8809 (7.42) & 0.8087 (8.26) & 0.8661 (7.06) \\
IS\textsubscript{ROS} & 0.8995 (5.74) & 0.8675 (5.87) & 0.8739 (5.82) \\
IS\textsubscript{SMOTE} & 0.9030 (5.30) & 0.8678 (5.75) & 0.8842 (5.45) \\
FS\textsubscript{RUS} & 0.9122 (4.17) & 0.8849 (4.12) & 0.9030 (4.24) \\
FS\textsubscript{ROS} & 0.9101 (4.27) & 0.8868 (3.77) & 0.8997 (4.30) \\
FS\textsubscript{SMOTE} & 0.9111 (4.15) & 0.8880 (3.77) & 0.8940 (4.17) \\
TS\textsubscript{SMOTE+SMOTE} & 0.9105 (4.10) & 0.8882 (3.60) & 0.8875 (4.08) \\
TS\textsubscript{SMOTE+RUS} & \textbf{0.9152 (3.70)} & \textbf{0.8898 (3.55)} & \textbf{0.9050 (3.73)} \\
\bottomrule
\end{tabularx}
\end{center}
\end{table*}

\section{Conclusions}
\label{sec:conc}

In this paper we considered the imbalanced image recognition problem with an application to the multi-class texture analysis in the colorectal cancer histology. We discussed the shortcomings of the existing data-level strategies of dealing with data imbalance in the context of image recognition and proposed a novel approach, two-stage data resampling, to mitigate the described deficiencies of over- and undersampling. Finally, in the conducted experimental analysis we empirically confirmed the usefulness of the proposed approach. We evaluated the negative impact of an artificially introduced data imbalance on the classification performance of a convolutional neural network, and examined the suitability of applying data resampling: directly in the image space, in the high-level feature space generated by a previously trained convolutional neural network, and a combination of the two. Proposed approach was motivated by, first of all, the necessity of preserving a large amount of training data, required for successful training of deep neural networks and limited by applying undersampling, and the need of limiting the distortions that are introduced to the data by applying synthetic oversampling strategies. Observed results indicate that, especially in the case of undersampling, a two-stage approach in which resampling is performed in the high-level feature space and leveraged to fine-tune the last layers of the network, leads to achieving a significantly better results than resampling directly in the image space. Furthermore, by applying a combination of oversampling the data in the image space and later undersampling it in the high-level feature space, we were able to achieve additional improvement in performance.

Due to the nature of the histological data, as well as the characteristics of the chosen benchmark dataset, further research is required to confirm the usefulness of the proposed approach in the general imbalanced image recognition setting. Specifically, evaluation on a large-scale dataset consisting of natural images with a higher dimensionality, such as ImageNet, would be beneficial to better understand the suitability of feature space resampling. This is due to the fact that image interpolation of natural images is more likely to introduce visual artifacts, making SMOTE interpolation directly in the image space less suitable, as well as the fact that an increased volume of data available for training could, in principle, make direct undersampling more suitable. Additionally, evaluation of different resampling strategies, as well as comparison with approaches other than data-level resampling algorithms, such as generative adversarial networks and training protocols utilizing imbalance-sensitive loss functions, would be beneficial for better understanding of the problem. Finally, another potentially viable research direction would be to consider additional methods for handling data imbalance in the initial stage that satisfy the original motivation or preserving the large amount of training data, such as using class-balanced loss functions. This is left for further research.
\section*{Acknowledgments}

This work was supported by the Polish National Science Center under the grant no. 2017/27/N/ST6/01705 as well as the PLGrid Infrastructure.

\bibliographystyle{IEEEtran}
\bibliography{bibliography}

\begin{thebibliography}{10}
\providecommand{\url}[1]{#1}
\csname url@samestyle\endcsname
\providecommand{\newblock}{\relax}
\providecommand{\bibinfo}[2]{#2}
\providecommand{\BIBentrySTDinterwordspacing}{\spaceskip=0pt\relax}
\providecommand{\BIBentryALTinterwordstretchfactor}{4}
\providecommand{\BIBentryALTinterwordspacing}{\spaceskip=\fontdimen2\font plus
\BIBentryALTinterwordstretchfactor\fontdimen3\font minus
  \fontdimen4\font\relax}
\providecommand{\BIBforeignlanguage}[2]{{%
\expandafter\ifx\csname l@#1\endcsname\relax
\typeout{** WARNING: IEEEtran.bst: No hyphenation pattern has been}%
\typeout{** loaded for the language `#1'. Using the pattern for}%
\typeout{** the default language instead.}%
\else
\language=\csname l@#1\endcsname
\fi
#2}}
\providecommand{\BIBdecl}{\relax}
\BIBdecl

\bibitem{chawla2002smote}
N.~V. Chawla, K.~W. Bowyer, L.~O. Hall, and W.~P. Kegelmeyer, ``{SMOTE}:
  synthetic minority over-sampling technique,'' \emph{Journal of artificial
  intelligence research}, vol.~16, pp. 321--357, 2002.

\bibitem{fernandez2018smote}
A.~Fern{\'a}ndez, S.~Garcia, F.~Herrera, and N.~V. Chawla, ``{SMOTE} for
  learning from imbalanced data: progress and challenges, marking the 15-year
  anniversary,'' \emph{Journal of artificial intelligence research}, vol.~61,
  pp. 863--905, 2018.

\bibitem{anand1993improved}
R.~Anand, K.~G. Mehrotra, C.~K. Mohan, and S.~Ranka, ``An improved algorithm
  for neural network classification of imbalanced training sets,'' \emph{IEEE
  Transactions on Neural Networks}, vol.~4, no.~6, pp. 962--969, 1993.

\bibitem{johnson2019survey}
J.~M. Johnson and T.~M. Khoshgoftaar, ``Survey on deep learning with class
  imbalance,'' \emph{Journal of Big Data}, vol.~6, no.~1, p.~27, 2019.

\bibitem{masko2015impact}
D.~Masko and P.~Hensman, ``The impact of imbalanced training data for
  convolutional neural networks,'' 2015.

\bibitem{lee2016plankton}
H.~Lee, M.~Park, and J.~Kim, ``Plankton classification on imbalanced large
  scale database via convolutional neural networks with transfer learning,'' in
  \emph{2016 IEEE international conference on image processing (ICIP)}.\hskip
  1em plus 0.5em minus 0.4em\relax IEEE, 2016, pp. 3713--3717.

\bibitem{pouyanfar2018dynamic}
S.~Pouyanfar, Y.~Tao, A.~Mohan, H.~Tian, A.~S. Kaseb, K.~Gauen, R.~Dailey,
  S.~Aghajanzadeh, Y.-H. Lu, S.-C. Chen \emph{et~al.}, ``Dynamic sampling in
  convolutional neural networks for imbalanced data classification,'' in
  \emph{2018 IEEE conference on multimedia information processing and retrieval
  (MIPR)}.\hskip 1em plus 0.5em minus 0.4em\relax IEEE, 2018, pp. 112--117.

\bibitem{buda2018systematic}
M.~Buda, A.~Maki, and M.~A. Mazurowski, ``A systematic study of the class
  imbalance problem in convolutional neural networks,'' \emph{Neural Networks},
  vol. 106, pp. 249--259, 2018.

\bibitem{koziarski2018convolutional}
M.~Koziarski, B.~Kwolek, and B.~Cyganek, ``Convolutional neural network-based
  classification of histopathological images affected by data imbalance,'' in
  \emph{Video Analytics. Face and Facial Expression Recognition}.\hskip 1em
  plus 0.5em minus 0.4em\relax Springer, 2018, pp. 1--11.

\bibitem{koziarski2019radial}
M.~Koziarski, ``Radial-{B}ased {U}ndersampling algorithm for classification of
  breast cancer histopathological images affected by data imbalance,'' in
  \emph{2019 12th International Congress on Image and Signal Processing,
  BioMedical Engineering and Informatics (CISP-BMEI)}.\hskip 1em plus 0.5em
  minus 0.4em\relax IEEE, 2019, pp. 1--5.

\bibitem{kwolek2019breast}
B.~Kwolek, M.~Koziarski, A.~Buka{\l}a, Z.~Antosz, B.~Olborski, P.~Wąsowicz,
  J.~Swad{\'z}ba, and B.~Cyganek, ``Breast cancer classification on
  histopathological images affected by data imbalance using active learning and
  deep convolutional neural network,'' in \emph{International Conference on
  Artificial Neural Networks}.\hskip 1em plus 0.5em minus 0.4em\relax Springer,
  2019, pp. 299--312.

\bibitem{mariani2018bagan}
G.~Mariani, F.~Scheidegger, R.~Istrate, C.~Bekas, and C.~Malossi, ``{BAGAN}:
  Data augmentation with balancing {GAN},'' \emph{arXiv preprint
  arXiv:1803.09655}, 2018.

\bibitem{cui2019class}
Y.~Cui, M.~Jia, T.-Y. Lin, Y.~Song, and S.~Belongie, ``Class-balanced loss
  based on effective number of samples,'' in \emph{Proceedings of the IEEE
  Conference on Computer Vision and Pattern Recognition}, 2019, pp. 9268--9277.

\bibitem{kather2016multi}
J.~N. Kather, C.-A. Weis, F.~Bianconi, S.~M. Melchers, L.~R. Schad, T.~Gaiser,
  A.~Marx, and F.~G. Z{\"o}llner, ``Multi-class texture analysis in colorectal
  cancer histology,'' \emph{Scientific reports}, vol.~6, p. 27988, 2016.

\bibitem{howard2017mobilenets}
A.~G. Howard, M.~Zhu, B.~Chen, D.~Kalenichenko, W.~Wang, T.~Weyand,
  M.~Andreetto, and H.~Adam, ``Mobile{N}ets: Efficient convolutional neural
  networks for mobile vision applications,'' \emph{arXiv preprint
  arXiv:1704.04861}, 2017.

\bibitem{deng2009imagenet}
J.~Deng, W.~Dong, R.~Socher, L.-J. Li, K.~Li, and L.~Fei-Fei, ``Image{N}et: A
  large-scale hierarchical image database,'' in \emph{2009 IEEE conference on
  computer vision and pattern recognition}.\hskip 1em plus 0.5em minus
  0.4em\relax Ieee, 2009, pp. 248--255.

\bibitem{chollet2015keras}
F.~Chollet \emph{et~al.}, ``Keras,'' \url{https://keras.io}, 2015.

\bibitem{lemaitre2017imbalanced}
G.~Lema{\^\i}tre, F.~Nogueira, and C.~K. Aridas, ``Imbalanced-learn: A python
  toolbox to tackle the curse of imbalanced datasets in machine learning,''
  \emph{The Journal of Machine Learning Research}, vol.~18, no.~1, pp.
  559--563, 2017.

\bibitem{Branco:2017}
P.~Branco, L.~Torgo, and R.~P. Ribeiro, ``Relevance-based evaluation metrics
  for multi-class imbalanced domains,'' in \emph{Advances in Knowledge
  Discovery and Data Mining - 21st Pacific-Asia Conference, {PAKDD} 2017, Jeju,
  South Korea, May 23-26, 2017, Proceedings, Part {I}}, 2017, pp. 698--710.

\end{thebibliography}

\end{document}